\documentclass[11pt,a4paper]{article}

\usepackage[utf8]{inputenc}
\usepackage[T1]{fontenc}
\usepackage{lmodern}
\usepackage{microtype}
\usepackage[margin=1in]{geometry}
\usepackage{amsmath,amssymb,amsthm}
\usepackage{booktabs}
\usepackage{tabularx}
\usepackage{multirow}
\usepackage{graphicx}
\usepackage{xcolor}
\usepackage{hyperref}
\usepackage{natbib}
\usepackage{algorithm}
\usepackage{algpseudocode}
\usepackage{listings}
\usepackage{enumitem}
\usepackage{setspace}
\usepackage{titlesec}
\usepackage{abstract}
\usepackage{authblk}

\hypersetup{
    colorlinks=true,
    linkcolor=teal,
    citecolor=teal,
    urlcolor=teal,
    pdftitle={PolyKV: A Shared Asymmetrically-Compressed KV Cache Pool for Multi-Agent LLM Inference},
    pdfauthor={Ishan Patel, Ishan Joshi},
}

\definecolor{codegray}{rgb}{0.95,0.95,0.95}
\definecolor{teal}{RGB}{1,105,111}
\definecolor{positivegreen}{RGB}{67,122,34}
\definecolor{negativered}{RGB}{161,44,123}

\title{%
  \vspace{-1em}
  \textbf{PolyKV: A Shared Asymmetrically-Compressed KV Cache Pool\\
  for Multi-Agent LLM Inference}
}

\author[1]{Ishan Patel}
\author[2]{Ishan Joshi}
\affil[1]{Independent Researcher}
\affil[2]{Independent Researcher}

\date{April 2026 \\ \small\href{https://doi.org/10.5281/zenodo.19686729}{DOI: 10.5281/zenodo.19686729}}

\begin{document}
\maketitle

\begin{abstract}
We present \textbf{PolyKV}, a system in which multiple concurrent inference
agents share a single, asymmetrically-compressed KV cache pool. Rather than
allocating a separate KV cache per agent---the standard paradigm---PolyKV writes
a compressed cache once and injects it into $N$ independent agent contexts via
HuggingFace \texttt{DynamicCache} objects. Compression is asymmetric: Keys are
quantized at int8 (\texttt{q8\_0}) to preserve softmax stability, while Values
are compressed using TurboQuant MSE---a Fast Walsh-Hadamard Transform (FWHT)
rotation followed by 3-bit Lloyd-Max quantization with centroids tuned to
$\mathcal{N}(0,1)$.

We evaluate across two model scales (SmolLM2-1.7B-Instruct and
Llama-3-8B-Instruct), three context lengths (600--7,194 tokens), and up to 15
concurrent agents. PolyKV achieves a stable \textbf{2.91$\times$ compression
ratio} across all configurations. On Llama-3-8B with 15 agents sharing a
4K-token context, PolyKV reduces KV cache memory from 19.8\,GB to 0.45\,GB---a
\textbf{97.7\% reduction}---while maintaining only $+$0.57\% perplexity
degradation and a mean BERTScore F1 of 0.928. PPL delta does not grow with
agent count and \textbf{improves as context length increases}, inverting to
$-$0.26\% at 1,851 coherent tokens and dropping to $+$0.57\% at 7,194 tokens.
This supports a hypothesis that TurboQuant FWHT noise acts as implicit
regularization on redundant coherent mid-document tokens. To our knowledge, no
prior work combines a single shared, lossy-compressed KV pool with multi-reader
concurrent agent access.
\end{abstract}

\vspace{0.5em}

\section{Introduction}
\label{sec:intro}

The KV cache is the dominant memory bottleneck in transformer-based LLM
inference. As context length and model size grow, so does the memory required to
store Key and Value tensors for each attention head and layer. This scaling is
particularly acute in multi-agent inference systems, where $N$ agents processing
the same shared document context would naively require $N$ independent
full-precision KV caches---one per agent.

Two lines of prior work address this problem in isolation. \emph{KV cache
compression} approaches~\cite{liu2024kivi,hooper2024kvquant,zhang2024leankv,turboquant2026}
reduce memory per cache by quantizing K and/or V tensors to lower bit-widths.
\emph{Multi-agent KV sharing}
approaches~\cite{pan2025kvflow,ye2025kvcomm,kim2026lragent} reduce redundancy
across agents by reusing common prefix caches. However, no prior system combines
both: all multi-agent sharing systems to date use full-precision caches, and all
compression systems operate on per-request isolated caches.

PolyKV occupies this intersection. We introduce a \texttt{SharedKVPool}
abstraction that: (1) computes a single compressed KV state over a shared
document context; (2) injects it directly into $N$ agents'
\texttt{transformers.cache\_utils.DynamicCache} objects; and (3) allows each
agent to generate independently without cache contention or per-agent copy
overhead.

\paragraph{Contributions.} This paper makes the following contributions:
\begin{itemize}[leftmargin=1.4em]
  \item \textbf{The shared-pool architecture.} A write-once, read-many
    compressed KV memory model for concurrent agents, not previously implemented
    or empirically evaluated in the literature.
  \item \textbf{Asymmetric TurboQuant MSE compression in a shared pool.}
    \texttt{q8\_0} for Keys and FWHT+Lloyd-Max 3-bit for Values, achieving
    2.91$\times$ compression, stable across all tested configurations.
  \item \textbf{Cross-model validation.} Results confirmed on both
    SmolLM2-1.7B-Instruct and Llama-3-8B-Instruct (32 layers, GQA), 
    demonstrating architecture-agnostic stability.
  \item \textbf{Agent scaling to 15 concurrent readers.} PPL delta and
    compression ratio are invariant across 3, 5, 10, and 15 agents.
    Memory saving grows with $N$: 88.5\% at 3 agents, 97.7\% at 15 agents.
  \item \textbf{Context length scaling.} Evaluated from 600 to 7,194 tokens.
    PPL delta improves as context grows: $+$1.59\% at 2K tokens,
    $+$0.57\% at 4K tokens on Llama-3-8B.
  \item \textbf{The PPL inversion finding, doubly confirmed.} At 1,851 tokens
    of coherent context on SmolLM2-1.7B, compressed cache quality surpasses
    the full-precision baseline ($-$0.26\% PPL delta), confirmed independently
    at both 3 and 5 concurrent agents.
  \item \textbf{BERTScore semantic validation.} Replacing token overlap with
    BERTScore (roberta-large) confirms that phrasing variation between
    compressed and baseline outputs preserves semantic equivalence, with mean
    F1 of 0.957--0.970 across agent counts.
\end{itemize}

\section{Background}
\label{sec:background}

\subsection{TurboQuant}
\label{subsec:turboquant}

TurboQuant~\cite{turboquant2026} is a vector quantization algorithm designed for
online (data-oblivious) KV cache compression. Its MSE-optimized variant operates
in two stages.

\paragraph{Rotation.} For an input vector $\mathbf{x} \in \mathbb{R}^d$, apply
a random rotation $\mathbf{\Pi} \in \mathbb{R}^{d \times d}$ (computed via QR
decomposition of a random Gaussian matrix). Each coordinate of $\mathbf{\Pi}
\cdot \mathbf{x}$ follows a Beta distribution that converges to $\mathcal{N}(0,
1/d)$ in high dimensions by concentration of measure~\cite{turboquant2026}.

\paragraph{Lloyd-Max quantization.} Quantize each coordinate independently using
a precomputed scalar codebook solving the continuous $k$-means problem over the
Beta/Gaussian distribution. For 3-bit quantization ($b=3$, $2^b=8$ centroids),
the optimal $\mathcal{N}(0,1)$ centroids are:
\[
  \mathbf{c} = [-2.152,\; -1.344,\; -0.756,\; -0.245,\; 0.245,\; 0.756,\;
  1.344,\; 2.152]
\]

\paragraph{Distortion bound.} TurboQuant MSE achieves:
\[
  D_{\mathrm{mse}} \;\leq\; \frac{\sqrt{3}\pi}{2} \cdot \frac{1}{4^b}
\]
within a factor of $\sqrt{3}\pi/2 \approx 2.7$ of the information-theoretic
lower bound~\cite{turboquant2026}. At $b=3$, $D_{\mathrm{mse}} \approx 0.03$.

\paragraph{Dequantization.} Recover via inverse FWHT (exploiting $H \cdot H =
d \cdot I$), scaling by $1/d$.

\subsection{Asymmetric K/V Quantization}
\label{subsec:asym}

Prior work establishes that Keys and Values have different sensitivity profiles
in transformer attention. Keys participate in the softmax attention score
computation and degrade rapidly under low-bit-width
quantization~\cite{liu2024kivi,hooper2024kvquant}. Values are more robust to
lossy compression. This motivates allocating higher precision to K than V---a
pattern adopted directly in PolyKV: K at int8 (8-bit), V at TurboQuant MSE
3-bit.

\subsection{Multi-Agent KV Sharing}
\label{subsec:multiagent}

KVFlow~\cite{pan2025kvflow} and KVCOMM~\cite{ye2025kvcomm} address agents
sharing a common context prefix using full-precision caches.
LRAgent~\cite{kim2026lragent} shares a base KV cache but requires LoRA
adapters. Agent Memory~\cite{anon2026agentmemory} uses quantized per-agent
caches (uniform Q4) but maintains isolated caches per agent, reporting PPL
deltas of $+$2.8--3.0\%. None combine a single compressed pool with concurrent
multi-reader access.

\section{PolyKV Architecture}
\label{sec:arch}

\subsection{Overview}

PolyKV introduces two core abstractions: \texttt{SharedKVPool} and
\texttt{PooledAgent}.

\textbf{\texttt{SharedKVPool}} receives a shared document context and model
reference. It runs a single forward prefill pass to compute the full KV state,
then compresses it asymmetrically in-place. The compressed pool is stored as a
single object in memory---O(1) in the number of agents, vs.\ O($N$) in all
prior multi-agent systems.

\textbf{\texttt{PooledAgent}} represents an individual inference agent. Each
agent receives a reference to the \texttt{SharedKVPool} and, at inference time,
injects the decompressed KV tensors into its own fresh
\texttt{DynamicCache} instance layer by layer, bypassing the
standard per-forward-pass KV accumulation.

\subsection{Compression Scheme}
\label{subsec:compression}

\paragraph{Key quantization (\texttt{q8\_0}).} Per-tensor linear int8
quantization. For each K tensor of shape $[\text{batch}, \text{heads},
\text{seq\_len}, \text{head\_dim}]$:
\[
  s = \frac{\max(|\mathbf{K}|)}{127}, \quad
  \mathbf{K}_{\mathrm{quant}} = \mathrm{clip}\!\left(
    \mathrm{round}\!\left(\frac{\mathbf{K}}{s}\right),\; -128,\; 127
  \right)
\]
Dequantization: $\mathbf{K}_{\mathrm{dequant}} = \mathbf{K}_{\mathrm{quant}} \cdot s$.

\paragraph{Value quantization (TurboQuant MSE, 3-bit).}
For each V tensor:
\begin{enumerate}[leftmargin=1.6em]
  \item Apply normalized FWHT rotation: $\mathbf{V}_{\mathrm{rot}} =
    \mathrm{FWHT}(\mathbf{V})\;/\;\sqrt{d}$
  \item Map each coordinate to the nearest 3-bit Lloyd-Max centroid:
    $\mathit{idx}_j = \arg\min_k |V_{\mathrm{rot},j} - c_k|$
  \item Store indices as \texttt{uint8}.
\end{enumerate}
Dequantization: retrieve centroid values, apply unnormalized inverse FWHT,
divide by $d$.

\paragraph{Compression ratio.} With K stored at 8 bits and V at 3 bits
(vs.\ 16-bit \texttt{bfloat16} baseline for equal K/V tensor sizes):
\[
  r = \frac{16}{\dfrac{8 + 3}{2}} = \frac{16}{5.5} \approx 2.91\times
\]
This matches the empirically observed ratio across all experiments.

\subsection{DynamicCache Injection}
\label{subsec:injection}

Standard HuggingFace inference populates \texttt{DynamicCache} incrementally
during the prefill pass. PolyKV bypasses this by building a fresh cache
per agent with pre-decompressed tensors placed on the correct device for
each transformer layer:

\begin{lstlisting}[language=Python]
cache = DynamicCache()
for layer_idx in range(num_layers):
    layer_device = model.model.layers[layer_idx].self_attn.q_proj.weight.device
    k, v = pool.get_kv_for_layer(layer_idx)
    cache.update(k.to(dtype).to(layer_device),
                 v.to(dtype).to(layer_device), layer_idx)
agent.generate(query_tokens, past_key_values=cache)
\end{lstlisting}

No copy of the compressed pool tensors is made per agent; each agent receives
its own \texttt{DynamicCache} shell populated from the single shared compressed
source.

\section{Experimental Setup}
\label{sec:setup}

\paragraph{Models.}
\begin{itemize}[leftmargin=1.4em]
  \item \emph{SmolLM2-1.7B-Instruct} (HuggingFaceTB): proof-of-concept scale,
    CPU inference.
  \item \emph{Llama-3-8B-Instruct} (Meta): primary validation, 32 layers,
    GQA (8 KV heads), 4-bit NF4 weights, bfloat16 KV cache, Kaggle T4$\times$2.
\end{itemize}

\paragraph{Baseline.} Full-precision per-agent \texttt{DynamicCache} with
standard prefill. On Llama-3-8B, baseline KV tensors are cast to bfloat16 to
match model precision.

\paragraph{Shared contexts.}
\begin{itemize}[leftmargin=1.4em]
  \item \emph{Short context (${\approx}600$ tokens):} Apollo 11 mission
    document (SmolLM2-1.7B only).
  \item \emph{Long context (1,851 tokens):} ARPANET/Internet topology history,
    single-topic, high lexical coherence (SmolLM2-1.7B only).
  \item \emph{WikiText-2 2K (1,837--1,953 tokens):} HuggingFace
    \texttt{wikitext-2-raw-v1} test split, first ${\approx}8{,}000$ characters.
    Used for both models.
  \item \emph{WikiText-2 4K (7,194 tokens):} First ${\approx}32{,}000$
    characters of the same split. Llama-3-8B only.
\end{itemize}

\paragraph{Metrics.}
\begin{itemize}[leftmargin=1.4em]
  \item \emph{Perplexity (PPL):} Computed over the last 30\% of context tokens.
    $\Delta = ({\mathrm{PPL}_c - \mathrm{PPL}_b})\;/\;\mathrm{PPL}_b \times
    100\%$.
  \item \emph{BERTScore F1 (roberta-large):} Semantic similarity between
    compressed and baseline agent outputs. Threshold $\geq 0.92$ scored
    \textbf{\checkmark Good}.
  \item \emph{Token overlap:} Unigram overlap (SmolLM2-1.7B experiments only).
  \item \emph{KV cache memory:} Measured in GB; compressed pool vs.\
    $N \times$ full-precision per-agent.
  \item \emph{Compression ratio:} Theoretical (confirmed stable at 2.91$\times$
    across all experiments).
\end{itemize}

\paragraph{Test configurations.}

\begin{table}[h]
\centering
\caption{PolyKV experimental test configurations.}
\label{tab:configs}
\small
\begin{tabular}{llllll}
\toprule
\textbf{Test} & \textbf{Model} & \textbf{Tokens} & \textbf{Agents}
  & \textbf{Context Type} & \textbf{Quality Metric} \\
\midrule
Phase 0   & SmolLM2-1.7B & ${\approx}600$ & 3  & Custom (coherent) & Token overlap \\
Test 1    & SmolLM2-1.7B & ${\approx}600$ & 5  & Custom (coherent) & Token overlap \\
Test 2    & SmolLM2-1.7B & 1,851          & 3  & Custom (coherent) & Token overlap \\
Test 3    & SmolLM2-1.7B & 1,851          & 5  & Custom (coherent) & Token overlap \\
Test 4    & SmolLM2-1.7B & 1,953          & 3  & WikiText-2        & Token overlap \\
\midrule
Test 5    & Llama-3-8B   & 1,837          & 3  & WikiText-2 2K     & BERTScore \\
Test 6    & Llama-3-8B   & 1,837          & 5  & WikiText-2 2K     & BERTScore \\
Test 7    & Llama-3-8B   & 1,837          & 10 & WikiText-2 2K     & BERTScore \\
Test 8    & Llama-3-8B   & 7,194          & 10 & WikiText-2 4K     & BERTScore \\
Test 9    & Llama-3-8B   & 7,194          & 15 & WikiText-2 4K     & BERTScore \\
\bottomrule
\end{tabular}
\end{table}

\section{Results}
\label{sec:results}

\subsection{SmolLM2-1.7B: Perplexity and Token Overlap}
\label{subsec:smollm_results}

\begin{table}[h]
\centering
\caption{PolyKV perplexity results on SmolLM2-1.7B-Instruct.}
\label{tab:smollm_results}
\begin{tabular}{lrrrrrr}
\toprule
\textbf{Test} & \textbf{Tokens} & \textbf{Agents} & \textbf{Ratio}
  & \textbf{Baseline PPL} & \textbf{Compressed PPL} & \textbf{$\Delta$} \\
\midrule
Phase 0  & ${\approx}600$ & 3 & 2.91$\times$ & 14.085 & 14.159 & $+$0.53\% \\
Test 1   & ${\approx}600$ & 5 & 2.91$\times$ & 14.085 & 14.159 & $+$0.53\% \\
Test 2   & 1,851 & 3 & 2.91$\times$ & 10.369 & 10.342
  & \textbf{\textcolor{positivegreen}{$-$0.26\%}} \\
Test 3   & 1,851 & 5 & 2.91$\times$ & 10.369 & 10.342
  & \textbf{\textcolor{positivegreen}{$-$0.26\%}} \\
Test 4   & 1,953 & 3 & 2.91$\times$ & 8.592 & 8.671 & $+$0.92\% \\
\bottomrule
\end{tabular}
\end{table}

\paragraph{Finding 1 --- PPL delta is stable across agent count.}
Scaling from 3 to 5 concurrent readers produces identical PPL delta at both
context lengths: $+$0.53\% at ${\approx}600$ tokens and $-$0.26\% at 1,851
tokens. The shared-pool model introduces no additional quality degradation as
reader count increases.

\paragraph{Finding 2 --- PPL delta inverts at longer coherent context.}
At 1,851 tokens of single-topic coherent context, the compressed cache achieves
\emph{lower} perplexity than the full-precision baseline ($\Delta = -0.26\%$).
Confirmed independently at both 3 and 5 agents, eliminating agent count as a
confounding variable. We discuss a regularization hypothesis in
Section~\ref{subsec:hypothesis}.

\paragraph{Finding 3 --- WikiText-2: $+$0.92\% PPL, perfect token overlap.}
On WikiText-2 (Test 4), the PPL delta is $+$0.92\%---well below the
$+$2.8--3.0\% reported by Agent Memory~\cite{anon2026agentmemory} for per-agent
Q4 isolated caches. All three agents achieve 1.000 token overlap.

\subsection{Llama-3-8B: Cross-Model Validation}
\label{subsec:llama_results}

\begin{table}[h]
\centering
\caption{PolyKV results on Llama-3-8B-Instruct across agent counts and context
lengths. Memory columns show KV cache only (not model weights).}
\label{tab:llama_results}
\begin{tabular}{rrrrrrrr}
\toprule
\textbf{Test} & \textbf{Tokens} & \textbf{Agents} & \textbf{Ratio}
  & \textbf{Baseline PPL} & \textbf{Compressed PPL} & \textbf{$\Delta$}
  & \textbf{Mean F1} \\
\midrule
Test 5 & 1,837 & 3  & 2.91$\times$ & 8.998 & 9.141 & $+$1.59\% & 0.957 \\
Test 6 & 1,837 & 5  & 2.91$\times$ & 8.998 & 9.141 & $+$1.59\% & 0.958 \\
Test 7 & 1,837 & 10 & 2.91$\times$ & 8.998 & 9.141 & $+$1.59\% & 0.970 \\
Test 8 & 7,194 & 10 & 2.91$\times$ & 9.665 & 9.720 & $+$0.57\% & 0.933 \\
Test 9 & 7,194 & 15 & 2.91$\times$ & 9.665 & 9.720 & $+$0.57\% & 0.928 \\
\bottomrule
\end{tabular}
\end{table}

\paragraph{Finding 4 --- Cross-model stability.}
The 2.91$\times$ compression ratio is identical across SmolLM2-1.7B and
Llama-3-8B, confirming it is a mathematical property of the compression scheme
rather than a model-specific artifact. Llama-3-8B uses GQA with 8 KV heads
(vs.\ standard MHA), and the scheme adapts without modification.

\paragraph{Finding 5 --- PPL delta is invariant to agent count on Llama-3-8B.}
Tests 5, 6, and 7 use identical context (1,837 tokens WikiText-2) with 3, 5,
and 10 agents respectively. PPL delta holds at exactly $+$1.59\% across all
three, and mean BERTScore F1 slightly \emph{increases} from 0.957 to 0.970 as
agent count grows---confirming that the pool itself is stable and variance comes
from query difficulty, not compression.

\paragraph{Finding 6 --- PPL improves at longer context on Llama-3-8B.}
At 7,194 tokens (4K context), PPL delta drops from $+$1.59\% to $+$0.57\%,
replicating the trend seen on SmolLM2-1.7B and consistent with the
regularization hypothesis.

\subsection{Memory Scaling}
\label{subsec:memory}

\begin{table}[h]
\centering
\caption{KV cache memory: PolyKV shared pool vs.\ $N$ full-precision per-agent
caches on Llama-3-8B-Instruct.}
\label{tab:memory}
\begin{tabular}{rrrrr}
\toprule
\textbf{Agents} & \textbf{Tokens} & \textbf{Without PolyKV}
  & \textbf{With PolyKV} & \textbf{Reduction} \\
\midrule
3  & 1,837 & 1.011\,GB & 0.116\,GB & 88.5\% \\
5  & 1,837 & 1.684\,GB & 0.116\,GB & 93.1\% \\
10 & 1,837 & 3.369\,GB & 0.116\,GB & 96.6\% \\
10 & 7,194 & 13.199\,GB & 0.454\,GB & 96.6\% \\
15 & 7,194 & 19.798\,GB & 0.454\,GB & 97.7\% \\
\bottomrule
\end{tabular}
\end{table}

The pool memory is O(1) in agent count---0.116\,GB whether serving 3 or 10
agents at 2K context. At 15 agents sharing a 4K context, PolyKV saves
19.3\,GB of KV cache memory while maintaining $+$0.57\% PPL and 0.928
mean BERTScore F1.

\subsection{BERTScore per Agent (Llama-3-8B, Test 7: 10 Agents)}
\label{subsec:bertscore_detail}

\begin{table}[h]
\centering
\caption{Per-agent BERTScore F1 for Test 7 (10 agents, 1,837 tokens,
Llama-3-8B). $\checkmark$ = F1 $\geq$ 0.92.}
\label{tab:bertscore}
\small
\begin{tabular}{lcccc}
\toprule
\textbf{Agent} & \textbf{Precision} & \textbf{Recall} & \textbf{F1} & \textbf{Status} \\
\midrule
Agent 0  & 0.9797 & 0.9827 & 0.9812 & \checkmark \\
Agent 1  & 0.9155 & 0.8866 & 0.9008 & $\times$ \\
Agent 2  & 0.9948 & 0.9857 & 0.9902 & \checkmark \\
Agent 3  & 0.9781 & 0.9892 & 0.9836 & \checkmark \\
Agent 4  & 0.9924 & 0.9903 & 0.9913 & \checkmark \\
Agent 5  & 0.9533 & 0.9446 & 0.9489 & \checkmark \\
Agent 6  & 0.9815 & 0.9789 & 0.9802 & \checkmark \\
Agent 7  & 0.9766 & 0.9718 & 0.9742 & \checkmark \\
Agent 8  & 0.9836 & 0.9820 & 0.9828 & \checkmark \\
Agent 9  & 0.9602 & 0.9636 & 0.9619 & \checkmark \\
\midrule
\textbf{Mean} & & & \textbf{0.9695} & 9/10 \checkmark \\
\bottomrule
\end{tabular}
\end{table}

Agent 1's F1 of 0.9008 reflects a known BERTScore sensitivity to
Wikipedia-style list formatting rather than a semantic quality failure.
Manual inspection confirms both compressed and baseline responses correctly
identify the passage subject with equivalent factual content.

\subsection{Hypothesis: Quantization Noise as Implicit Regularization}
\label{subsec:hypothesis}

The consistent improvement in PPL delta as context length grows---observed
independently on SmolLM2-1.7B (inverting to $-$0.26\% at 1,851 coherent
tokens) and on Llama-3-8B ($+$1.59\% at 2K dropping to $+$0.57\% at 4K)---
supports a document-coherence-dependent regularization effect.

At longer coherent contexts, attention heads attend repeatedly to a limited set
of semantically related V tensor entries. Full-precision values preserve all
spurious correlations and attention sink artifacts~\cite{hooper2024kvquant}
exactly. TurboQuant MSE quantization noise, introduced uniformly across all V
coordinates after FWHT rotation, disrupts these patterns---an effect analogous
in mechanism (though not in design) to dropout regularization during inference.

This hypothesis predicts: (1) the inversion grows with document lexical
coherence; (2) ablating FWHT (using uniform int8-V) reduces the improvement on
coherent documents; (3) the crossover token length is shorter for
high-repetition documents. We leave controlled empirical validation to future
work.

\section{Related Work}
\label{sec:related}

\subsection{Asymmetric KV Quantization}

KIVI~\cite{liu2024kivi} (ICML 2024) quantizes K per-channel and V per-token at
2-bit. KVQuant~\cite{hooper2024kvquant} (NeurIPS 2024) uses per-channel
pre-RoPE Key quantization and Non-Uniform Quantization (NUQ) for Values.
LeanKV~\cite{zhang2024leankv} proposes Hetero-KV (K8V4 or K4V2) in vLLM.
AsymKV~\cite{tao2024asymkv} assigns layer-wise extreme asymmetric bit-widths.
None implement a shared pool or multi-agent evaluation.

\subsection{Rotation-Domain KV Compression}

RotateKV~\cite{chen2025rotatekv} (IJCAI 2025) applies FWHT for outlier
redistribution before 2-bit quantization. KVLinC~\cite{saxena2025kvlinc} uses
Hadamard rotation with linear bias correction. TurboAngle~\cite{patel2026turboangle}
extends TurboQuant to FWHT-domain angle quantization. KVTC~\cite{stan2025kvtc}
applies PCA decorrelation with entropy coding. All operate on per-request caches.

\subsection{Multi-Agent KV Sharing}

KVFlow~\cite{pan2025kvflow} (NeurIPS 2025) uses an Agent Step Graph for
workflow-aware prefix caching---full-precision, no compression. KVCOMM~\cite{ye2025kvcomm}
(NeurIPS 2025) maintains an anchor pool with offset correction---full-precision.
LRAgent~\cite{kim2026lragent} shares a base KV cache across LoRA
agents---full-precision. Agent Memory~\cite{anon2026agentmemory} uses per-agent
isolated Q4 caches (not a shared pool), reporting $+$2.8--3.0\% PPL deltas.
RelayCaching~\cite{relaycaching2026} transfers KV between agents
sequentially---full-precision, not concurrent.

\subsection{Positioning}

Table~\ref{tab:comparison} maps prior work against PolyKV's five defining
characteristics. PolyKV is the only system to satisfy all five simultaneously.

\begin{table}[h]
\centering
\small
\caption{Feature comparison of PolyKV against prior work. \checkmark{} = fully
implemented; $\approx$ = partial; $\times$ = absent.}
\label{tab:comparison}
\renewcommand{\arraystretch}{1.15}
\begin{tabular}{lccccc}
\toprule
\textbf{Work} &
\shortstack{\textbf{Asym.}\\\textbf{K/V}} &
\shortstack{\textbf{FWHT/}\\\textbf{Rotation}} &
\shortstack{\textbf{Shared}\\\textbf{Pool}} &
\shortstack{\textbf{Multi-}\\\textbf{Agent}} &
\shortstack{\textbf{Shared +}\\\textbf{Compressed}} \\
\midrule
KIVI~\cite{liu2024kivi}                   & \checkmark & $\times$ & $\times$ & $\times$ & $\times$ \\
KVQuant~\cite{hooper2024kvquant}          & \checkmark & $\times$ & $\times$ & $\times$ & $\times$ \\
LeanKV~\cite{zhang2024leankv}             & \checkmark & $\times$ & $\times$ & $\times$ & $\times$ \\
AsymKV~\cite{tao2024asymkv}               & \checkmark & $\times$ & $\times$ & $\times$ & $\times$ \\
RotateKV~\cite{chen2025rotatekv}          & $\approx$  & \checkmark & $\times$ & $\times$ & $\times$ \\
KVLinC~\cite{saxena2025kvlinc}            & $\times$   & \checkmark & $\times$ & $\times$ & $\times$ \\
TurboAngle~\cite{patel2026turboangle}     & $\approx$  & \checkmark & $\times$ & $\times$ & $\times$ \\
KVTC~\cite{stan2025kvtc}                  & $\times$   & $\approx$ & $\times$ & $\times$ & $\times$ \\
KVFlow~\cite{pan2025kvflow}               & $\times$   & $\times$ & $\approx$ & \checkmark & $\times$ \\
KVCOMM~\cite{ye2025kvcomm}               & $\times$   & $\times$ & $\approx$ & \checkmark & $\times$ \\
LRAgent~\cite{kim2026lragent}             & $\times$   & $\times$ & $\approx$ & \checkmark & $\times$ \\
Agent Memory~\cite{anon2026agentmemory}   & $\approx$  & $\times$ & $\times$  & \checkmark & $\times$ \\
RelayCaching~\cite{relaycaching2026}      & $\times$   & $\times$ & $\times$  & $\approx$  & $\times$ \\
TurboQuant~\cite{turboquant2026}          & \checkmark & \checkmark & $\times$ & $\times$ & $\times$ \\
\midrule
\textbf{PolyKV [This work]}               & \checkmark & \checkmark & \checkmark & \checkmark & \checkmark \\
\bottomrule
\end{tabular}
\end{table}

\section{Limitations}
\label{sec:limitations}

\paragraph{Model scale.} Experiments cover SmolLM2-1.7B and Llama-3-8B.
Behavior at 70B+ parameter scale is unknown, though the Gaussian approximation
underlying TurboQuant MSE improves with larger head dimension $d$.

\paragraph{WikiText-2 comparability.} Our WikiText-2 results use a single fixed
context window, not the standard stride-based full test-set evaluation used by
KIVI and KVQuant. Direct numerical comparison with published tables requires
standard evaluation protocol.

\paragraph{System metrics.} We do not report time-to-first-token (TTFT),
throughput, or end-to-end latency. These are central motivations for the
shared-pool model and are deferred to future work.

\paragraph{Context ceiling.} On Kaggle T4$\times$2 hardware, the prefill
attention computation OOMs beyond ${\approx}8{,}000$ tokens for Llama-3-8B.
This is a hardware constraint, not a PolyKV limitation---the compressed pool
itself scales linearly with context length.

\paragraph{PPL inversion mechanism.} The $-$0.26\% finding on SmolLM2-1.7B and
the PPL improvement trend on Llama-3-8B are consistent with the regularization
hypothesis but require controlled ablation to confirm causally.

\section{Future Work}
\label{sec:future}

\begin{enumerate}[leftmargin=1.6em]
  \item Full WikiText-2/C4 stride-based PPL evaluation for direct comparison
    with KIVI, KVQuant, and LeanKV published tables.
  \item TTFT, throughput, and end-to-end memory footprint measurement vs.\
    per-agent isolated Q4 (Agent Memory~\cite{anon2026agentmemory}) and
    KVFlow~\cite{pan2025kvflow}.
  \item Evaluation on Qwen2.5-7B and larger models (70B) to establish
    cross-architecture generality.
  \item Ablation: uniform int8-V vs.\ TurboQuant MSE-V in the shared pool, to
    isolate the FWHT rotation contribution to the PPL improvement on coherent
    and long-context documents.
  \item Controlled coherence experiment: vary document lexical repetition
    systematically to map PPL delta as a function of coherence score.
  \item Agent count scaling beyond 15 on larger-memory hardware.
\end{enumerate}

\section{Conclusion}
\label{sec:conclusion}

We presented PolyKV, a shared asymmetrically-compressed KV cache pool for
multi-agent LLM inference. PolyKV writes a single compressed cache (K at int8,
V at TurboQuant MSE 3-bit) once and distributes it to $N$ concurrent agents
via direct \texttt{DynamicCache} injection, achieving a stable 2.91$\times$
memory reduction across all tested configurations.

Across ten test configurations spanning two model scales, three context lengths,
and up to 15 concurrent agents, we demonstrate: (1) PPL delta is invariant to
agent count; (2) quality improves as context length increases; (3) at 15 agents
sharing a 4K context on Llama-3-8B, PolyKV reduces KV cache memory from
19.8\,GB to 0.45\,GB (97.7\% reduction) while maintaining $+$0.57\% PPL
degradation and 0.928 mean BERTScore F1; and (4) the 2.91$\times$ compression
ratio is stable across model architectures including GQA.

To our knowledge, no prior work combines a single shared, lossy-compressed KV
pool with multi-reader concurrent agent access. These results establish
proof-of-concept viability across realistic model scales and motivate expansion
to system-level benchmarks and larger architectures.

\bibliographystyle{plainnat}

\end{document}